\documentclass[final]{cvpr}

\usepackage{times}
\usepackage{epsfig}
\usepackage{graphicx}
\usepackage{amsmath}
\usepackage{amssymb}
\usepackage{algorithm}
\usepackage{algorithmicx}
\usepackage{algpseudocode}
\usepackage{amsmath}

\usepackage{subfigure}
\usepackage{booktabs,bigstrut}
\usepackage[american]{babel}
\usepackage{microtype}
\usepackage[pagebackref=true,breaklinks=true,colorlinks,bookmarks=false]{hyperref}

\def\cvprPaperID{3819} 
\def\confYear{CVPR 2021}

\begin{document}
	
	\title{Domain Adaptive Object Detection via Feature Separation and Alignment}
	
\author{Chengyang Liang$^*\!$, Zixiang Zhao\thanks{Equal contribution.}\ , Junmin Liu$^\dagger\!$, Jiangshe Zhang\thanks{Corresponding authors.}\\
	Xi’an Jiaotong University\\
	Xi'an, Shaanxi, P.R. China\\
	{\tt\small \{liangcy1996,zixiangzhao\}@stu.xjtu.edu.cn}\\
	{\tt\small \{junminliu,jszhang\}@mail.xjtu.edu.cn}
}

	\maketitle

\begin{abstract}
	Recently, adversarial-based domain adaptive object detection (DAOD) methods have been developed rapidly. However, there are two issues that need to be resolved urgently. Firstly, numerous methods reduce the distributional shifts only by aligning all the feature between the source and target domain, while ignoring the private information of each domain. Secondly, DAOD should consider the feature alignment on object existing regions in images. 
	But redundancy of the region proposals and background noise could reduce the domain transferability.
	Therefore, we establish a Feature Separation and Alignment Network (FSANet) which consists of a gray-scale feature separation (GSFS) module, a local-global feature alignment (LGFA) module and a region-instance-level alignment (RILA) module.
	The GSFS module decomposes the distractive/shared information which is useless/useful for detection by a dual-stream framework, to focus on intrinsic feature of objects and resolve the first issue. 
	Then, LGFA and RILA modules reduce the distributional shifts of the multi-level features.
	Notably, scale-space filtering is exploited to implement adaptive searching for regions to be aligned, and instance-level features in each region are refined to reduce redundancy and noise mentioned in the second issue.
	Various experiments on multiple benchmark datasets prove that our FSANet achieves better performance on the target domain detection and surpasses the state-of-the-art methods.
\end{abstract}

\section{Introduction}

Object detection, as a hot topic of computer vision, is to find out the objects of interest on the image and to determine their locations and categories.
In the era of deep learning~\cite{NIPS2012_4824}, many deep convolutional neural network (DNN)-based methods~\cite{girshick2015fast,ren2015faster,Redmon_2016_CVPR,Redmon_2017_CVPR,redmon2018yolov3,liu2016ssd} have been proposed to make the detection models with excellent performance in natural image benchmark datasets~\cite{everingham2010pascal, lin2014microsoft}.
While in reality, there is an obvious distributional gap between the training set (source domain) and the testing set (target domain), which may lead to a significant reduction in the generalization ability of models in target domain. 

Therefore, many unsupervised domain adaptive object detection~(DAOD) methods~\cite{DAfaster, saito2019strong, Xinge2019Adapting, kim2019diversify, cai2019exploring, Chen2020Harmonizing} are proposed to solve the problem of domain distributional shifts.
Most unsupervised DAOD methods are nested adversarial training module within advanced object detection frameworks, such as Faster R-CNN~\cite{ren2015faster}. By minimizing the domain shifts of multi-level feature maps, domain adaption is achieved to improve the detection accuracy in the target domain.
Numerous DAOD methods are proposed and most of them focus on image-level~\cite{DAfaster}, instance-level~\cite{DAfaster,Chen2020Harmonizing}, pixel-level~\cite{kim2019diversify, Chen2020Harmonizing}, region-level~\cite{Xinge2019Adapting} and strong-local \& weak-global-level~\cite{saito2019strong} alignments.
Whereas, the existing multi-level alignment modules may cause three issues:	
Firstly, they only focus on the mapping of the source and target domain, and less consider other distractive information that is unnecessary to be aligned, \eg, background information.
Secondly, we should concern over the regions where the object needs to be aligned on the image, and ignore background noise that is useless for detection.
However, the current region searching algorithm \cite{Xinge2019Adapting} in region-level alignment 
restrict the category numbers of candidate region. Since there is no label for the target domain, we cannot exactly determine the object region numbers.
Lastly, for the existing instance alignment modules, 
they are all based on ROI-Pooling~\cite{ren2015faster} feature alignment. While the training of region proposal network (RPN)~\cite{ren2015faster} mainly depends on the labels of the source domain, and there may be abundant redundancy or noise in region of interests (ROIs) on the target domain. Resultly, when the instance is aligned, there will be an extra alignment of background features or multiple alignments of instance-level features for some certain objects.

In order to solve the above mentioned issues, we propose a DAOD method based on the architecture of Faster R-CNN~\cite{ren2015faster}, called feature separation and alignment network (FSANet). The network consists of a gray-scale feature separation (GSFS) module, a local-global feature alignment module (LGFA) and a region-instance-level alignment (RILA) module. The feature separation module separates the distractive information containing the features that do not need to be aligned and the shared information related to the object detection through the difference loss, so that the feature alignment module can concentrate on the object information. LGFA and RILA are domain adaptation components, which reduce the distributional gaps of the multi-level features. Our contributions can be divided into two-fold:

(1) The dual-stream network~\cite{bousmalis2016domain, zhaoijcai2020} can extract multi-model information of different domain images. 	
Inspired by this, we design a dual-stream auto-encoder network for the GSFS. The distractive information of the image is separated by a private encoder, and the object detection backbone network acts as a shared information extractor to obtain useful features for detection.
When aligning high-dimensional features, we use shared features instead of the whole image features to reduce the impact of distractive information that is not related to detection, so as to improve domain adaptability.
Particularly, in order to reduce the impact of image color differences and the difficulty of reconstruction, we use the gray-scale image to extract the distractive features. This operation can make the private encoder focus on detection-related/irrelevant features of the objects rather than the image color, and while simplifying the reconstruction task.

(2) We design the RAIL module, which can effectively solve the negative effects caused by the redundancy of the region proposals and background noise, and determine the category number for candidate region adaptively. Initially, by exploiting Scale-Space Filtering algorithm (SSF)~\cite{895974}, we cluster the center coordinate of the bounding boxes predicted by RPN, to implement an adaptive region selection. Thus the traditional clustering region number uncertainty problem is well solved.
Then, after obtaining the cluster centers and scale by SSF, the outliers of ROIs can be found. We believe that these abnormal ROIs have a high probability of containing background information. So we exclude them before region grouping, thereby reducing the effect of ROIs noise. Finally, through global pooling layer, the instance-level features of each group (\ie, candidate region) are refined and instance alignment is performed to figure out the problem of ROIs redundancy.

In summary, the adaptability and transferability of adversarial-based adaptive detection methods are enhanced by separating the shared and distractive information of the source/target domain and aligning the region-instance-level features.
Extensive experiments illustrate that the proposed FSANet has reached an outstanding level of cross-domain detection performance on multiple benchmark datasets.
For example, the FSANet achieves $42.7\%$ mAP for transfer task on PASCAL $\to$ Clipart1k, which surpassing the state-of-the-art (SOTA) adversarial-based adaptive methods~\cite{DAfaster,Xinge2019Adapting,saito2019strong,kim2019diversify,Chen2020Harmonizing,sindagi2019prior,he2020domain}.


\section{Related Work}
\noindent
{\bf Object Detection.} 
Nowadays, with the rapid development of deep neural convolutional networks (DNN), lots of methods have been proposed to solve the object detection task. They can be divided into two-stage object detection methods based on region proposals and end-to-end one-stage methods.
The pioneering work of the two-stage methods is R-CNN~\cite{girshick2014rich}, which first utilizes a selective search method~\cite{uijlings2013selective} to extract region proposals from images, and then trains a network to classify each ROIs. SPP-Net~\cite{he2015spatial} and Fast R-CNN~\cite{girshick2015fast} exploited DNNs to propose region proposals and significantly improved  detection accuracy and speed. As DNNs were extended to share convolutional feature maps among all ROIs~\cite{ren2015faster} , the end-to-end two-stage methods (\eg, Faster R-CNN~\cite{ren2015faster} and RetinaNet~\cite{lin2017focal}) were proposed.
On the other hand, typical methods of one-stage object detection algorithms are YOLO~\cite{Redmon_2016_CVPR, Redmon_2017_CVPR, redmon2018yolov3} and SSD~\cite{liu2016ssd}, which extract the feature maps from the original image based on the convolutional network, and then directly perform object classification and bounding box regression. \medskip\\
{\bf Domain Adaptive Object Detection.} The existing unsupervised DAOD methods can be divided into three categories~\cite{li2020deep}: adversarial-based, reconstruction-based and hybrid adaptive methods.
For the adversarial-based adaptive methods,
Chen \etal~\cite{DAfaster} first applied the unsupervised domain adaptive method to object detection. By combining with the gradient reversal layer and domain classifier, an adversarial-based method~\cite{ganin2015unsupervised} is proposed to solve the domain shift in the real scene. At the same time, it is pointed out that the core issue of unsupervised DAOD is to solve the domain gaps in image and instance level.
For example, Zhu \etal~\cite{Xinge2019Adapting} indicated that the object detection needs to concern with the object rather than the entire image features, and then a regional-level domain adaptive network based on  GAN~\cite{goodfellow2014generative} is established to solve the region proposal redundancy problem. 
Saito \etal~\cite{saito2019strong} proposed a local domain classifier network based on FCN~\cite{Long2017Fully} to achieve the alignment of low-level features. 
For the reconstruction-based adaptive methods,
Arruda \etal~\cite{Arruda2019Cross} firstly generated pseudo samples similar to the target domain from the source samples by CycleGAN~\cite{Zhu2017Unpaired}, and added them to the training set to improve detection accuracy.
The hybrid adaptive methods aim at combining the above two categories to improve the model performance.
In order to realize the adaptative object detection from natural images to artistic images, Inoue \etal~\cite{Inoue2018Cross} initially trained a detection model from the source domain samples, then fine-tuned the model on the pseudo samples of source domain, and finally implemented the weak training on the target domain samples.
Based on pseudo samples, Kim \etal~\cite{kim2019diversify} proposed an adaptive method for domain diversification and multi-domain invariant representation.
Chen \etal~\cite{Chen2020Harmonizing} added the attention mechanism to the network based on~\cite{saito2019strong}. 
Although the above methods have achieved good performance, they do not concern for separating the distractive information and solving the region proposals redundancy and background noise. So in our paper, we propose a novel DAOD method called FSANet to further figure out the above mentioned issues.
\begin{figure*}[t]
	\begin{center}
		\includegraphics[width=0.77\linewidth]{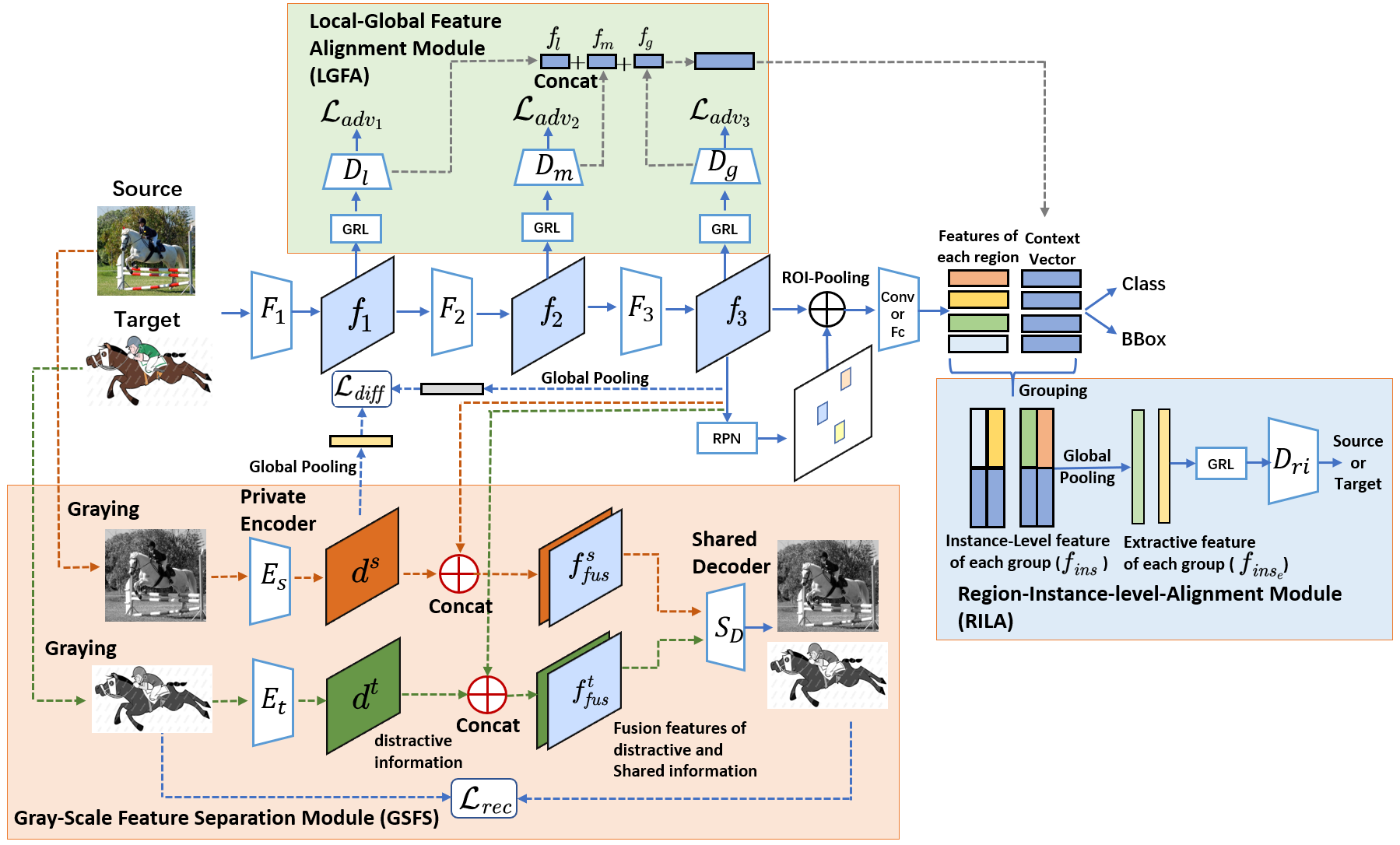}
	\end{center}
	\caption{The overall structure of the proposed FSANet.}
	\label{fig:net}
\end{figure*}
\section{Method}
In this section, we introduce the details of FSANet. The framework is illustrate in Figure~\ref{fig:net}, which consists of four parts, namely the \textit{object detection} module, the \textit{LGFA} module, the \textit{RILA} module and the \textit{GSFS} module. 	
Initially, the paired source/target domain RGB images are input to the detection network, and the detection loss is calculated with the label of the source domain images, and the instance-level features alignment is realized by the RILA module. For the multi-level features $f_{\tau}(\tau=1,2,3)$ extracted by the backbone network $F_{\tau}$, they are input into the LGFA module to complete the multi-level features alignment. In addition, the gray-scale paired images are input into the GSFS module, and the useful/useless features for detection are separated through a specific loss.
Here, we use Faster R-CNN~\cite{ren2015faster} as the object detection backbone architecture, and the LGFA module follows the design of~\cite{saito2019strong,Chen2020Harmonizing}.	
Especially, we focus on the GSFS module and the RILA module.

For the notations in this paper, the source domain dataset with labeled samples are denoted by $D_s=\{(x_i^s, b_i^s, y_i^s)\}_{i}^{n_s}$, where $x_i^s$ is the input source image, $b_i^s$ is the coordinates of the bounding box, $y_i^s$ is the object category label, and $n_s$ is the number of samples in the source domain dataset. Meanwhile, unlabeled target domain dataset is represented as $D_t=\{(x_i^t)\}_i^{n_t}$, where $n_t$ is the number of samples in the target dataset. Our task is to transfer the model learned from $D_s$ to $D_t$.

\subsection{Gray-Scale Feature Separation Module}
As is known to all, domain adversarial loss can reduce the inconsistency between the features distribution of the two domains. However, there will still be some distractive information that cannot be reduced, \eg, the background distractive information of the source/target domain.
So we propose a GSFS module based on a dual-stream network, and employ it to separate the distractive/shared information which is useless/useful for object detection with the high-level features through the private encoder $E_s$ and $E_t$, the shared decoder $S_D$ and the features extraction module $F_\tau$, as shown in Figure~\ref{fig:net}. Firstly, $F_3$ extracts high-level features related to object detection via the detection loss. Then domain adversarial loss of LGFA module makes the features $f_\tau$ of the source/target domain aligned. As a result, $f_3$ is the shared high-level features useful for detection.
For the GSFS module, we define the different loss to limit the similarity of $\left\lbrace d^s,f_3^s\right\rbrace $ and $\left\lbrace d^t,f_3^t\right\rbrace$:
\begin{equation}
\mathcal{L}\!_\mathit{diff}\!=\!\frac{1}{n_s}\!\sum_{i=1}^{n_s} \lVert g(d^s_i)^\top\!g(f_{3,i}^s) \rVert _F^2 + \frac{1}{n_t}\!\sum_{i=1}^{n_t}\!\lVert g(d^t_i)^ \top\!g(f_{3,i}^t) \rVert _F^2,
\end{equation}
where $\lVert \cdot \rVert _F^2$ is the squared Frobenius norm , $g(\cdot)$ is the global pooling layer, $d_i^s$, $d_i^t$, $f_{3,i}^s$, $f_{3,i}^t$ denote distractive information output by private encoder $E_s$/$E_t$ and the shared information $f_3$ of source/target domain for $i$th image, respectively.

In addition, considering that simple restrictions will make $d^s$ and $d^t$ meaningless, and the information related to objects cannot include all the information of the input image, we use the fusion features $f_\mathit{fus}^s\!=\![d^s, f_{3}^s]$, $f_\mathit{fus}^t\!=\![d^t, f_{3}^t]$ as the input of the shared decoder $S_D$ to reconstruct the input image, where $[\cdot,\cdot]$ denotes the concatenation of two feature maps in the channel dimension. Through the restriction of image reconstruction, the separated features can possess almost all the information of source/target images. So that the private encoder extracts the distractive information irrelevant to detection in the image.
Notably, the private encoder inputs two-domain images in gray-scale, and extract the distractive information which is irrelevant to the color of the two-domain images.
If the input is an RGB image, it will not only increase the difficulty of reconstruction, but also make the private encoder disturb by the color difference of the image, while ignoring other distractive information on the two data domains.

Finally, the reconstruction loss of the gray-scale images is defined as:
\begin{equation}
\mathcal{L}_\mathit{rec} = \frac{1}{n_s}\sum_{i=1}^{n_s} \lVert x_{g,i}^s - \hat{x}_{g,i}^s \rVert _1^1 +
\frac{1}{n_t}\sum_{i=1}^{n_t} \lVert x_{g,i}^t - \hat{x}_{g,i}^t \rVert _1^1,
\end{equation}
where $\lVert \cdot \rVert _1^1$ is the $\ell_1$-norm, ${x}_{g,i}^s$, ${x}_{g,i}^t$, $\hat{x}_{g,i}^s$ and $\hat{x}_{g,i}^t$ are the $i$th gray-scale input images and reconstructed images by the shared decoder $S_D$ in source/target domain. 

\subsection{Region-Instance-Level Alignment}
The RILA module consists of a \textit{grouping} component and a \textit{context-aware RILA} component. Please refer to supplemental material to see illustration and more details.\\
\noindent
{\bf Grouping.} Some alignment module~\cite{DAfaster, he2019multi, Chen2020Harmonizing} implement the instance-level feature alignment of all region proposals. However, the absence of labels on the target domain leads to the training of RPN mainly guided by the labeled data on the source domain~\cite{ren2015faster}. Therefore, we cannot ensure that the region proposals on the target domain contain the objects, where may contain a lot of background noise.
Meanwhile, objects are generally detected by multiple region proposals, which means that some region proposals are redundant. 
We should perform feature alignments for the instance-level features in the region proposals where the object is most likely to exist, instead of aligning all features.
As we all know, the locations of region proposals where contain the same object at multiple scales should be similar if they are predicted by a reliable RPN.
A natural idea is to cluster the similar region proposal coordinates to reduce redundancy for ROIs and eliminate the background noises by excluding outliers.

In particular, we use RPN to get the predicted bounding box $\{b_{i,x},b_{i,y},w_i,h_i\}_{i=1}^N$ of the region proposals, where $(b_{i,x},b_{i,y})$ is the center coordinate of the $i$th predicted bounding box, $w_i$ is its width, and $h_i$ is its height. Then we plan to exploit the scale-space filtering~(SSF)~\cite{895974} algorithm to cluster the center coordinates to adaptively obtain $K$ cluster centers, which means that ROIs can be divided into $K$ categories, and thereby, the instance-level features can be indirectly divided  into $K$ categories. As a result, we only need to align the instance-level features in each category. \\
\noindent
{\bf Adaptive candidate region searching.}
SSF is a clustering method that adaptive to the category numbers without a manual setting. It is very suitable for the grouping module due to the unknown object numbers in the target domain. Therefore, inspired by~\cite{895974}, we decided to use SSF to improve the traditional K-means algorithm to accomplish the clustering of bounding boxes.
By regarding each sample as a ``light point'', the SSF algorithm formulates the clustering issue as finding the ``lighting blob center'' under different blur scales, \ie, different scale-space maps. The specific workflow can be divided into the following steps:\\
\indent
(1) \textit{Clustering center iteration.} For a given dataset  $\mathcal{X}=\left\{x_{i} \in \mathbb{R}^{2}: i=1, \ldots, N\right\}$ (in this paper, this is a set of center coordinates of predicted bounding boxes by RPN), we define $P(x, \sigma_j)$ as the scale-space map for $\mathcal{X}$ under the blur scale $\sigma_j$. It can be calculated by $P(x, \sigma_j)=p(x) * \phi(x, \sigma_j)$, where $p(x)$ means the scatter image of data samples in $\mathcal{X}$, $\phi(\cdot, \sigma_j)$ is the Gaussian blur kernel with the blur scale $\sigma_j$, and $*$ represents the convolution operator. So the set of clustering centers $\mathcal{C}(\sigma_j)$ under different scales $\sigma_{j}$ is obtained by $\nabla_{x} P(x, \sigma_j)=0$.
In iteration steps, $\sigma_j$ is firstly updated, then after $\mathcal{C}(\sigma_j)$ has converged, a larger $\sigma_{j+1}$ is obtained by $\sigma_{j+1}\!=\!k\sigma_{j}$ ($k$ is a hyperparameter and $k\!>\!1$), and  $\mathcal{C}(\sigma_{j+1})$ are iterated again until all samples belong to one category. More details can be seen in the supplemental material.

(2) \textit{Adaptive category number selection.} With the completion of the above iterations, we have obtained a serious of clustering center, \ie, $\{\mathcal{C}(\sigma_1), \cdots,\mathcal{C}(\sigma_j),\cdots,\mathcal{C}(\sigma_J)\}$. We hope to select the clustering model that best fit for the data distribution. So the ``lifetime''~\cite{895974} is used as a measure of the stability for category number, which is calculated by
\begin{equation}\label{eqs:pi}
\pi(\sigma_j)=c \log (\sigma_j / \varepsilon),
\end{equation}
where $\varepsilon=0.01, c=1 / \log (1.05)$.
Notably, the interval $ \mathbf{\Sigma}=\left\lbrace \sigma_{J_{inf}}, \sigma_{J_{inf}+1},\dots,\sigma_{J_{sup}} \right\rbrace $  with the longest ``lifetime'' represent the relative appropriate scales.
Finally, the most appropriate clustering model $\{\mathcal{C}(\sigma^*),\sigma^*\}$ equals to median of $\mathbf{\Sigma}$, and the adaptive category number $K$ equals to the element number of $\mathcal{C}(\sigma^*)$.

(3) \textit{Remove outliers.}
After obtaining the optimal $\{\mathcal{C}(\sigma^*),\sigma^*\}$, we can use the model to group the centers $\{(b_{i,x},b_{i,y})\}_{i=1}^N$ of the predicted bounding box of all region proposals. 
For each bounding box center $b_i=(b_{i,x},b_{i,y})$, if $\lVert b_i-c_k^* \rVert _2>\sigma^*$, we consider $b_i$ to be an outlier, and its corresponding region proposal probably contains the background noise, which should be excluded from grouping.
Through this method, we can restrict certain outlier region proposals from participating in instance-level alignment, and find the instance features that need to be aligned in the image more robustly.\\
\noindent
{\bf Context-Aware Region-Instance-Level Alignment.}
After clustering the proposals, we need to refine the instance-level features of each category for instance alignment.
For simplicity, we reshape the features of instances belonging to the same category into the same size and then concatenate them into a feature map as the input of the domain classifier.
But the effect of redundant or noisy region proposals is not fundamentally solved.
Therefore, we use the global pooling layer to refine the instance-level features in the same category. 
Let $\Theta_k\in \mathbb{R}^{m_k\times d}$ denote the instance-level features of the $k$th category obtained by SSF, where $m_k$ is the number of the $k$th category , $d$ is the dimension of the instance-level features.
Through the global pooling layer, we can get the feature of the $k$th category $\hat{\Theta}_k\in \mathbb{R}^{1\times d}$, which can fully reflect the instance-level features of the $k$ category and thereby reduce the influence of redundancy and noise.

Then, we use the context fusion instance-level features~\cite{saito2019strong} as the input of bounding box regression and classification prediction in the instance alignment and object detection module. We acquire three different levels of context vectors $f_l, f_m, f_g$ from the domain classifier $D_l, D_m, D_g$, and get the instance-level features $f_r$ of each region proposals based on ROI-Pooling~\cite{he2017mask}. Through feature concatenation, the fused context instance-level features $f_{ins}=[f_l,f_m,f_g,f_r]$ is input into the global pooling layer and obtain refined feature of each category $f_{ins_e}$, as shown in Figure~\ref{fig:net}.

Subsequently, we input the global pooling instance-level features into the domain classifier $G_{ri}$, and output the domain labels to achieve regional instance alignment. For the loss function of this module, we use Focal Loss~\cite{lin2017focal,saito2019strong} as follows:
\begin{equation*}
\mathcal{L}_\mathit{ri_s}\!=\!-\frac{1}{n_s} \sum_{i=1}^{n_s} \frac{1}{K_i^s}\sum_{k=1}^{K_i^s}(1\!-\!\mathit{D}_\mathit{ri}(\mathit{f}_\mathit{ins_e,i,k}^{s}))^\gamma\log(\mathit{D}_\mathit{ri}(\mathit{f}_\mathit{ins_e,i,k}^{s})),
\end{equation*}
\begin{equation*}
\mathcal{L}_\mathit{ri_t}\!=\!-\frac{1}{n_t} \sum_{i=1}^{n_t} \frac{1}{K_i^t}\sum_{k=1}^{K_i^t}(\mathit{D}_\mathit{ri}(\mathit{f}_\mathit{ins_e,i,k}^{t}))^\gamma\log(1\!-\!\mathit{D}_\mathit{ri}(\mathit{f}_\mathit{ins_e,i,k}^{t})),
\end{equation*}

\begin{equation}
\mathcal{L}_{ri}=\frac{1}{2}(\mathcal{L}_{ri_s}+\mathcal{L}_{ri_t}),
\label{kmeansloss}
\end{equation}
where $K_i^s$ is the number of adaptive categories for $i$th source sample, $f_\mathit{ins_e,k}^{s}$ is the instance-level features of the $k$th region of the source domain after global pooling, and $\gamma$ is the parameter of Focal Loss.

In short, this module can solve the redundancy of the region proposals and the influence of the background noise, so that the instance alignment is focused on the features where the object is existing.

\subsection{Traing Loss}
The loss of object detection includes the classification loss $\mathcal{L}_{c}$ on the source domain and the regression loss $\mathcal{L}_{r}$ of the bounding boxes~\cite{ren2015faster}.
For the LGFA module, we use the Local Feature Masks and IWAT-I frameworks mentioned in~\cite{Chen2020Harmonizing}, and the corresponding adversarial loss is $\mathcal{L}_{lg}\!=\!\mathcal{L}_{adv_1}\!+\!\mathcal{L}_{adv_2}\!+\!\mathcal{L}_{adv_3}$ in Figure~\ref{fig:net}.
By combining all modules mentioned above, the overall objective function of the model is
\begin{equation}\label{loss}
\max_{D} \min_{F, E, S} \mathcal{L}_{c} + \mathcal{L}_{r} + 
\beta(\mathcal{L}_{rec} + \mathcal{L}_{diff}) - 
\lambda(\mathcal{L}_{lg} + \mathcal{L}_{ri}),	
\end{equation}
where $D,F,E,S$ denote the domain classifier, object detection framework, private encoder and shared decoder, respectively. $\lambda$ and $\beta$ are turning parameters. The sign of gradients for the last item is flipped by a gradient reversal layer proposed by~\cite{ganin2015unsupervised}.

\section{Experiments}
In this section, we show the effectiveness of FSANet in adaptive object detection experiments on three benchmark domain shifts datasets including real $\to$ artistic, normal $\to$ foggy and synthetic $\to$ real, \ie, PASCAL~\cite{everingham2010pascal} to Clipart1k~\cite{Inoue2018Cross}, Cityscapes~\cite{cordts2016cityscapes} to Foggy-Cityscapes~\cite{sakaridis2018semantic} and Sim10k~\cite{johnson2016driving} to Cityscapes.

\begin{table*}[t]
	\setlength{\tabcolsep}{1.5pt}
	\begin{center}
		\resizebox{\textwidth}{!}{
			\begin{tabular}{lccccccccccccccccccccc}
				\toprule 
				Methods     & aero  & bike & bird & boat & bot & bus & car & cat & chair & cow & table & dog & horse & mbike & persn & plant & sheep & sofa & train & tv & mAP \\
				\midrule
				Source Only &  35.6    & 52.5          & 24.3          & 23.0          & 20.0  & 43.9  & 32.8  & 10.7  & 30.6  & 11.7  & 13.8  & 6.0   & 36.8  & 45.9  & 48.7  & 41.9  & 16.5  & 7.3   & 22.9  & 32.0  & 27.8  \\
				SWDA \cite{saito2019strong}    & 26.2          & 48.5          & 32.6          & 33.7          & 38.5  & 54.3  & 37.1  & 18.6  & 34.8  & 58.3  & 17.0  & 12.5  & 33.8  & 65.5  & 61.6  & \textbf{52.0}  & 9.3   & 24.9  & 54.1  & 49.1  & 38.1  \\
				HTCN \cite{Chen2020Harmonizing} & 33.6          & 58.9          & 34.0          & 23.4          & 45.6  & 57.0  & 39.8  & 12.0  & 39.7  & 51.3  & 21.1  & 20.1  & 39.1  & 72.8  & 63.0  & 43.1  & 19.3  & 30.1  & 50.2  & 51.8  & 40.3  \\
				DDMRL \cite{kim2019diversify}    & 25.8          & 63.2          & 24.5          & \textbf{42.4} & \textbf{47.9}  & 43.1  & 37.5  & 9.1   & \textbf{47.0}  & 46.7  & 26.8  & 24.9  & \textbf{48.1}  & 78.7  & 63.0  & 45.0  & 21.3  & \textbf{36.1}  & 52.3  & \textbf{53.4}  & 41.8  \\
				ATF \cite{he2020domain}  & \textbf{41.9}          & 67.0          & 27.4          & 36.4          & 41.0  & 48.5  & 42.0  & 13.1  & 39.2  & \textbf{75.1}  & \textbf{33.4}  & 7.9   & 41.2  & 56.2  & 61.4  & 50.6  & \textbf{42.0}  & 25.0  & 53.1  & 39.1  & 42.1  \\
				\midrule
				FSANet   & 31.0          & 63.7          & \textbf{34.8} & 29.4  & 43.0  & \textbf{70.7}  & 40.8  & 18.7  & 39.6  & 57.4  & 22.2  & \textbf{27.0}  & 33.3  & \textbf{85.6}  & 63.3  & 45.7  & 21.9  & 24.7  & \textbf{56.7}  & 44.5  & \textbf{42.7}  \\
				\bottomrule
			\end{tabular}
		}
	\end{center}
	\caption{Results on adpatation from PASCAL VOC to Clipart Dataset (real $\to$ artistic). Source only stands for Faster R-CNN that is trained only using source domain without adaptation. Average precision ($\%$) is evaluated on target images. The backbone network is ResNet-101.}
	\label{Clipart}
\end{table*}

\subsection{Datasets}
\noindent
{\bf PASCAL $\to$ Clipart1k.} We conduct domain adaptive experiments from real images to artistic images to show that FSANet is effective in dissimilar domains, where PASCAL VOC dataset~\cite{everingham2010pascal} and Clipart1k~\cite{Inoue2018Cross} are set as the real source domain and the target artistic domain, respectively. The PASCAL dataset contains 20 categories of images and their bounding boxes. In this experiment, the training and validation splits of PASCAL VOC 2007 and 2012 are used for training, resulting in about 15K images.  Clipart1k has a total of 1K images, which contains the same class as PASCAL. All images in it are employed for training (w/o labels) and testing. \\
{\bf Cityscape $\to$ Foggy-Cityscapes.} For experiments in similar domains, we use Cityscapes and Foggy-Cityscapes as the source and target domain dataset, respectively. 
The Cityscapes~\cite{cordts2016cityscapes} dataset contains street scenes in different cities under normal weather conditions captured by a vehicle-mounted camera. There are 2975 images in the training set and 500 images in the test set. The label data are acquired by~\cite{DAfaster}. In this experiment, we employed the training set of Cityscapes for training. Foggy-Cityscapes~\cite{sakaridis2018semantic} is obtained by adding fog noise to the Cityscapes dataset, and its label is the same as Cityscapes. 
We set the training set of Foggy-Cityscapes for training and the test set for calculating average accuracy. \\
{\bf Sim10k $\to$ Cityscapes.} In the experiments from synthetic images to real images, we conduct the experiment for Sim10k $\to$ Cityscapes. Sim10k~\cite{johnson2016driving} is acquired in a game scene of a computer game Grand Theft Auto V. It has 10k computer-synthesized driving scene images. According to the protocol of~\cite{DAfaster}, we only evaluated detection accuracy on cars. All images of Sim10k are set for training.

\subsection{Implementation Details}
In all experiments, the object detection model follows the settings of~\cite{saito2019strong,Xinge2019Adapting,Chen2020Harmonizing}, using Faster R-CNN~\cite{ren2015faster} with ROI-alignment~\cite{he2017mask}, where the shorter side of the images equal to 600 and the hyper-parameters in the object detection are set based on the setting of~\cite{ren2015faster}. For different dataset, the backbone network utilize VGG-16~\cite{simonyan2014very} or ResNet-101~\cite{he2016deep} with the parameter initialization weights following the pre-trained on ImageNet~\cite{deng2009imagenet}. A stochastic gradient descent training model with a momentum of 0.9 is used, and the learning rate for the first 50K iterations is set to 0.001 and then decays to 0.0001. In each iteration, we input a pair of source and target domain images. After 70K iterations, we calculate the mean average precision (mAP) where the IoU threshold is 0.5. We set $\gamma\!=\!5$ in Eq.~(\ref{kmeansloss}) and $\beta\!=\!0.1$ in Eq.~(\ref{loss}) in all experiments. $\lambda$ is set to 1 for PASCAL $\to$ Clipart1k and Cityscapes $\to$ Foggy-Cityscapes while set to 0.1 for Sim10k $\to$ Cityscapes. Our experiments are implemented by the Pytorch~\cite{paszke2017automatic}.

\subsection{Performance Comparison}
We compare our FASNet with various SOTA DAOD methods, including Domain adaptive Faster-RCNN ({DA-Faster}) \cite{DAfaster}, Selective Cross-Domain Alignment ({SCDA} )~\cite{Xinge2019Adapting}, Strong-Weak Distribution Alignment ({SWDA})~\cite{saito2019strong}, Domain Diversification and Multi-domain-invariant Representation Learning ({DDMRL})~\cite{kim2019diversify}, Hierarchical Transferability Calibration Network ({HTCN})~\cite{Chen2020Harmonizing}, Prior-based Domain Adaptive Object Detection ({PBDA}~\cite{sindagi2019prior}) and Asymmetric Tri-way Faster-RCNN ({ATF})~\cite{he2020domain}. 
We cite the quantitative results in their original papers for comparison.

\noindent
{\bf Results on real $\to$ artistic. } The performance of FASNet in the target domain is significantly better than all comparison methods, as displayed in Table~\ref{Clipart}. Compared with SOTAs, mAP is improved by $+0.6\%$ (from $42.1\%$ to $42.7\%$), which fully illustrates that the proposed method can improve the transferability for real $\to$ artistic.

\noindent
{\bf Results on normal $\to$ foggy.} According to Table~\ref{cstocs_fg}, FASNet achieves the best performance and approaches the upper bound of the transfer task on Cityscape $\to$ Foggy-Cityscapes.

\noindent
{\bf Results on synthetic $\to$ real.} As shown in Table~\ref{sim2cs}, Our FSANet exceeds all comparison methods in the adaptive task on Sim10k $\to$ Cityscape, which further demonstrates that our method has better transferability and adaptivity. 
\begin{table}[t]
	\small
	\setlength{\tabcolsep}{1pt}
	\begin{center}
		\begin{tabular}{lccccccccc}
			\toprule	
			Method      & persn & rider & car & truck & bus & train & mbike & bcycle & mAP \\
			\midrule
			Source Only  & 24.1             & 33.1          & 34.3          & 4.1           & 22.3  & 3.0   & 15.3  & 26.5  & 20.3  \\
			DA-Faster \cite{DAfaster}       & 25.0          & 31.0          & 40.5          & 22.1  & 35.3  & 20.2  & 20.0  & 27.1  & 27.6  \\
			SCDA \cite{Xinge2019Adapting}   & 33.5          & 38.0          & 48.5          & 26.5  & 39.0  & 23.3  & 28.0  & 33.6  & 33.8  \\
			SWDA \cite{saito2019strong}     & 29.9          & 42.3          & 43.5          & 24.5  & 36.2  & 32.6  & 30.0  & 35.3  & 34.3  \\
			DDMRL \cite{kim2019diversify}      & 30.8          & 40.5          & 44.3          & 27.2  & 38.4  & 34.5  & 28.4  & 32.2  & 34.6  \\
			ATF \cite{he2020domain}         & 34.6          & 47.0          & 50.0          & 23.7  & 43.3  & 38.7  & 33.4  & \textbf{38.8} & 38.7 \\
			PBDA \cite{sindagi2019prior}    & \textbf{36.4} & 47.3          & \textbf{51.7} & 22.8  & 47.6  & 34.1  & \textbf{36.0} & 38.7  & 39.3 \\
			HTCN \cite{Chen2020Harmonizing} & 33.2          & 47.5          & 47.9  & 31.6  & 47.4  & \textbf{40.9} & 32.3  & 37.1  & 39.8 \\
			\midrule
			FSANet      & 33.5          & \textbf{47.6} & 45.8          & \textbf{32.0} & \textbf{50.1} & 37.7  & 35.7  & 37.7  & \textbf{40.0} \\
			\midrule
			Oracle      & 33.2          & 45.9  & 49.7  & 35.6          & 50.0  & 37.4  & 34.7  & 36.2  & 40.3 \\
			\bottomrule
		\end{tabular}
	\end{center}
	\caption{Results on adpatation from Cityscapes to Foggy-Cityscapes (normal $\to$ foggy). The backbone network is VGG-16.}
	\label{cstocs_fg}
\end{table}
\begin{table}[t]
	\begin{center}
		\begin{tabular}{lclc}
			\toprule
			Method                      &   AP on car   & Method                          & AP on car \\ 
			\midrule
			Source Only                 &     34.6      & DA-Faster \cite{DAfaster}       &   38.9    \\
			SWDA \cite{saito2019strong} &     40.1      & HTCN \cite{Chen2020Harmonizing} &   42.5    \\
			ATF \cite{he2020domain}     &     42.8      & SCDA \cite{Xinge2019Adapting}   &   43.0    \\
			\midrule
			FSANet                      & \textbf{43.2} &                                 &           \\ 
			\bottomrule
		\end{tabular}
	\end{center}
	\caption{Results on adpatation from Sim10k to Cityscapes (synthetic $\to$ real). The backbone network is VGG-16.}
	\label{sim2cs}
\end{table}
\begin{figure*}[t]
	\begin{center}
		\subfigure{\includegraphics[width=0.3\linewidth]{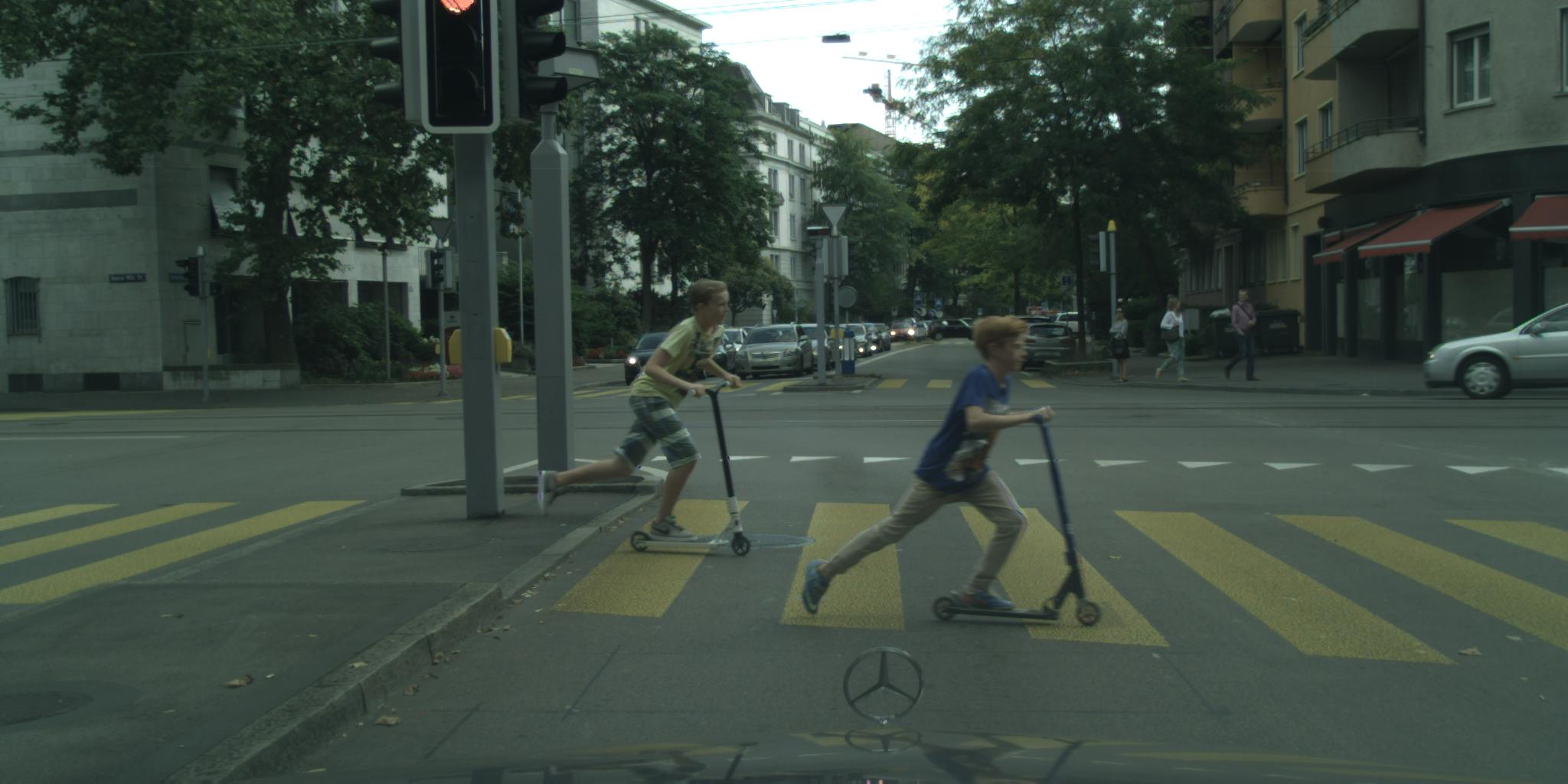}}
		\subfigure{\includegraphics[width=0.3\linewidth]{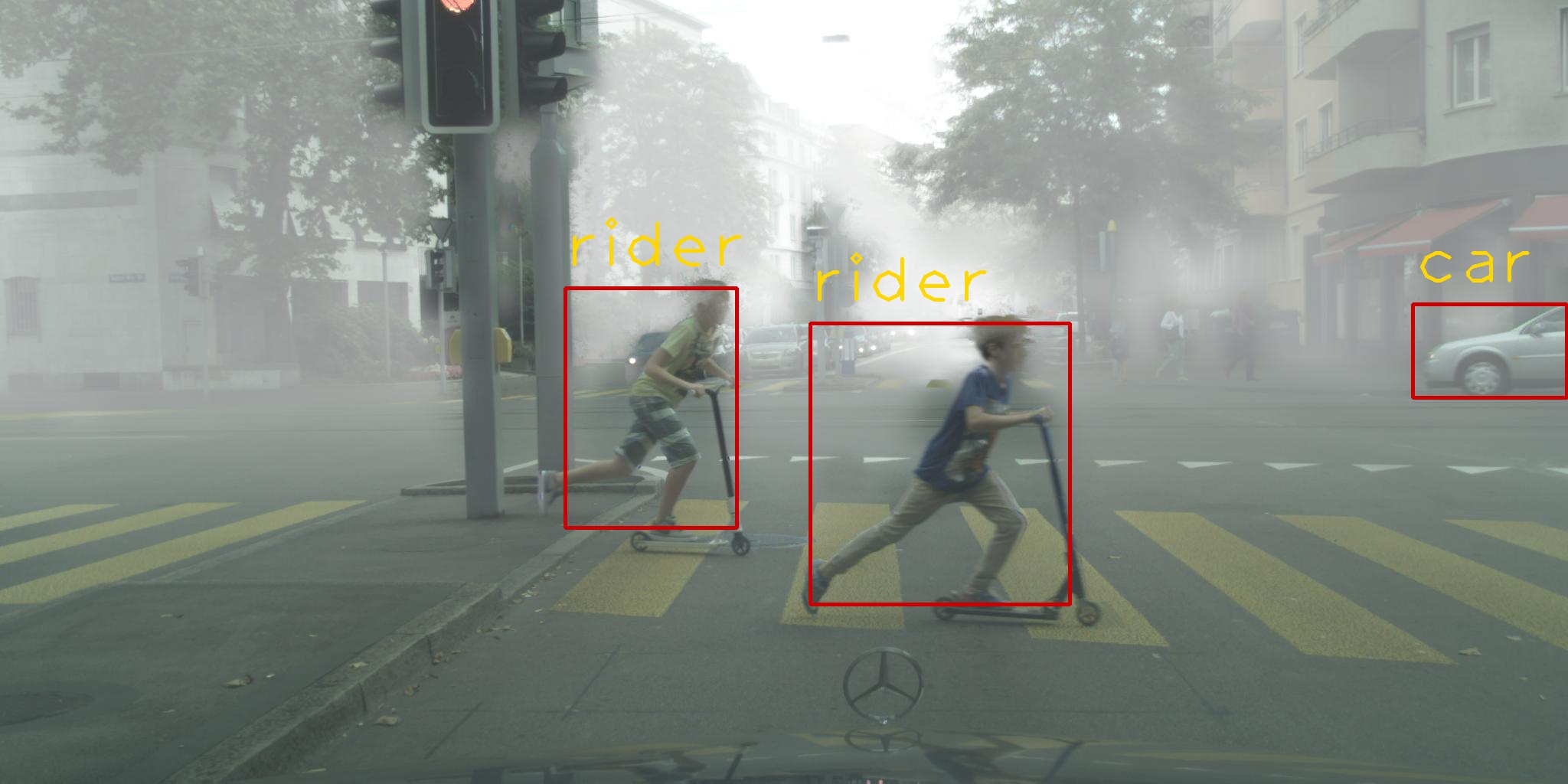}}
		\subfigure{\includegraphics[width=0.3\linewidth]{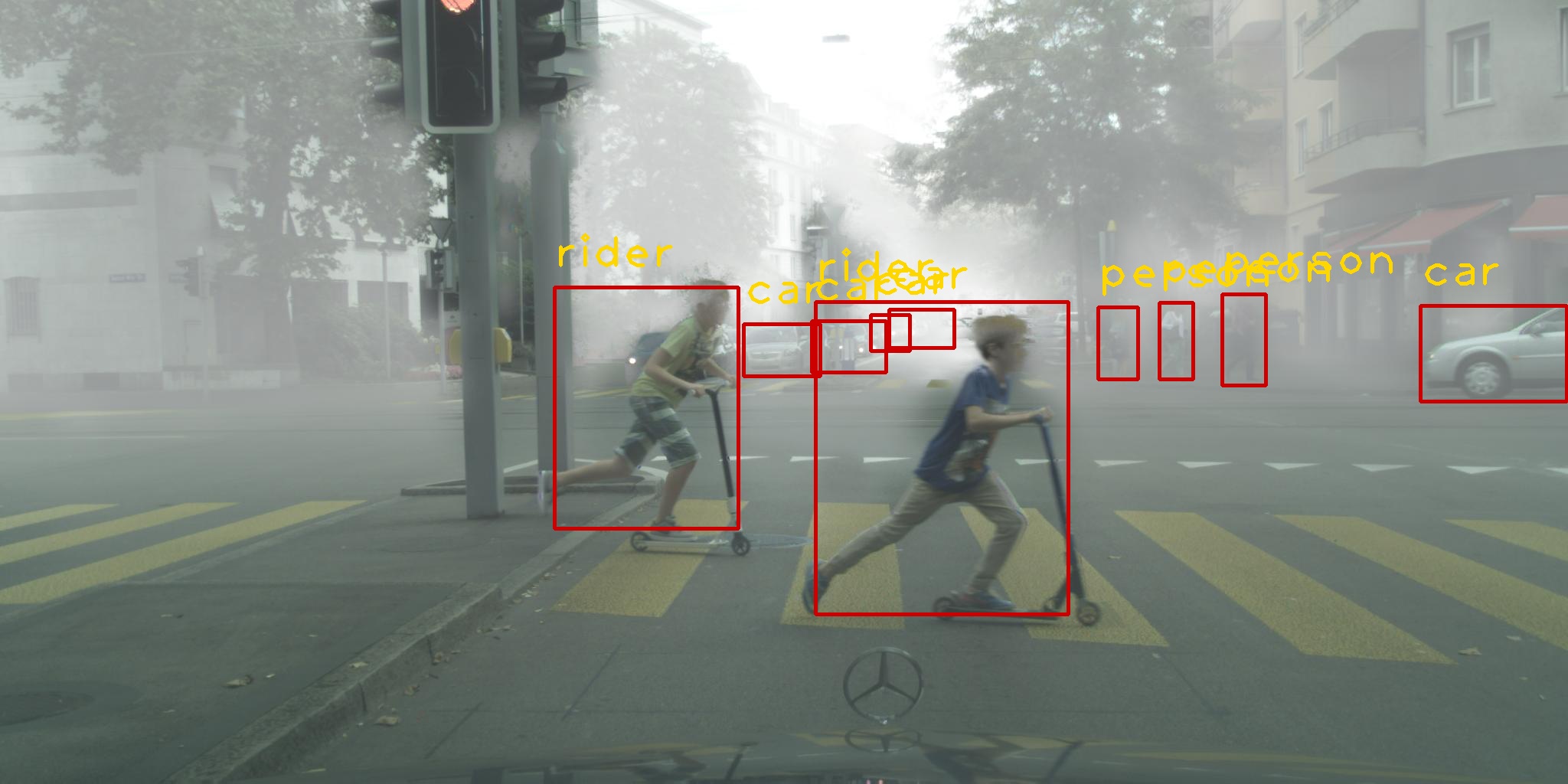}}\vspace{-2.5mm}\\
		\setcounter{subfigure}{0}
		\subfigure[Original Image]{\includegraphics[width=0.3\linewidth]{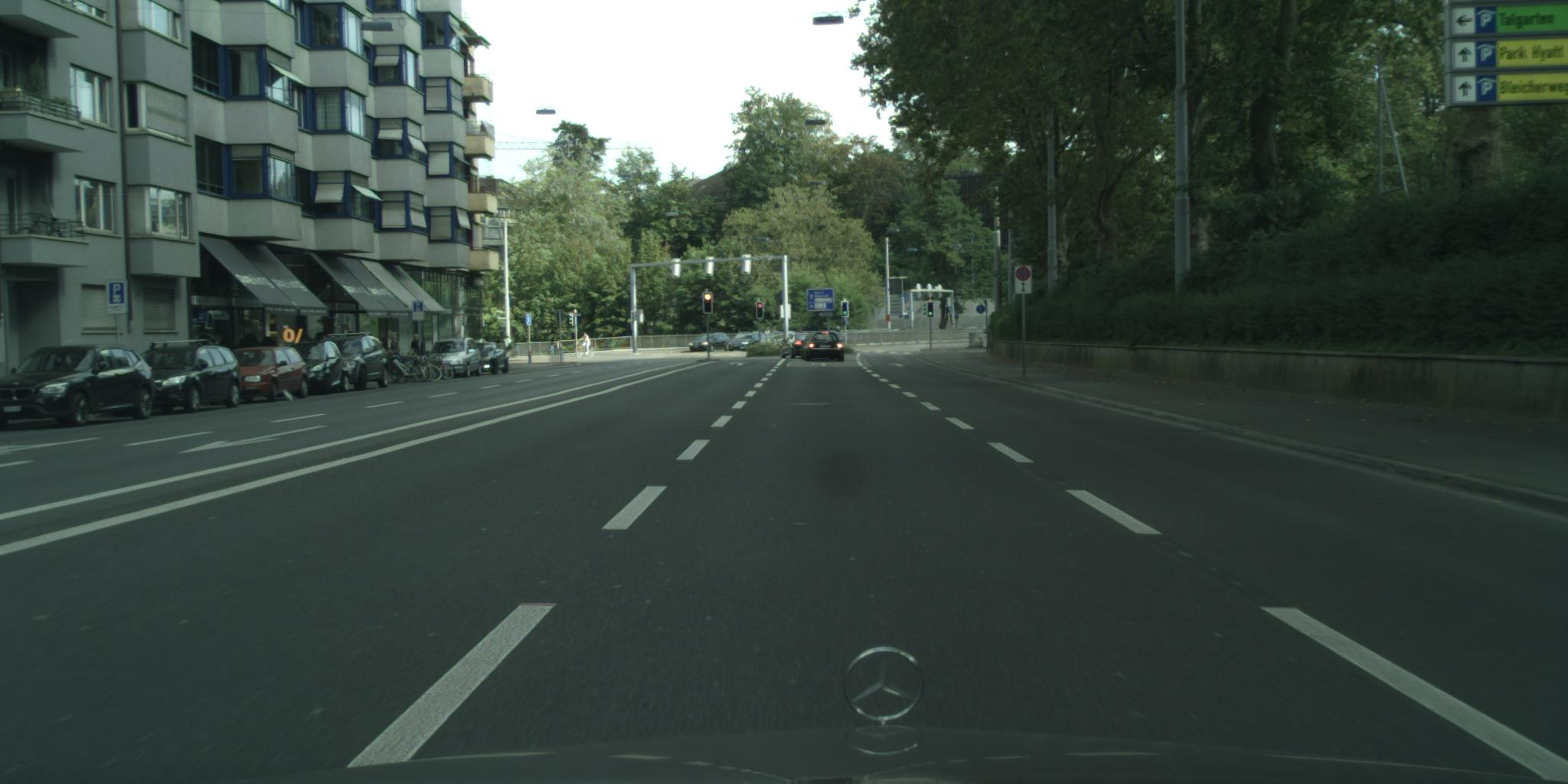}}
		\subfigure[Source Only]{\includegraphics[width=0.3\linewidth]{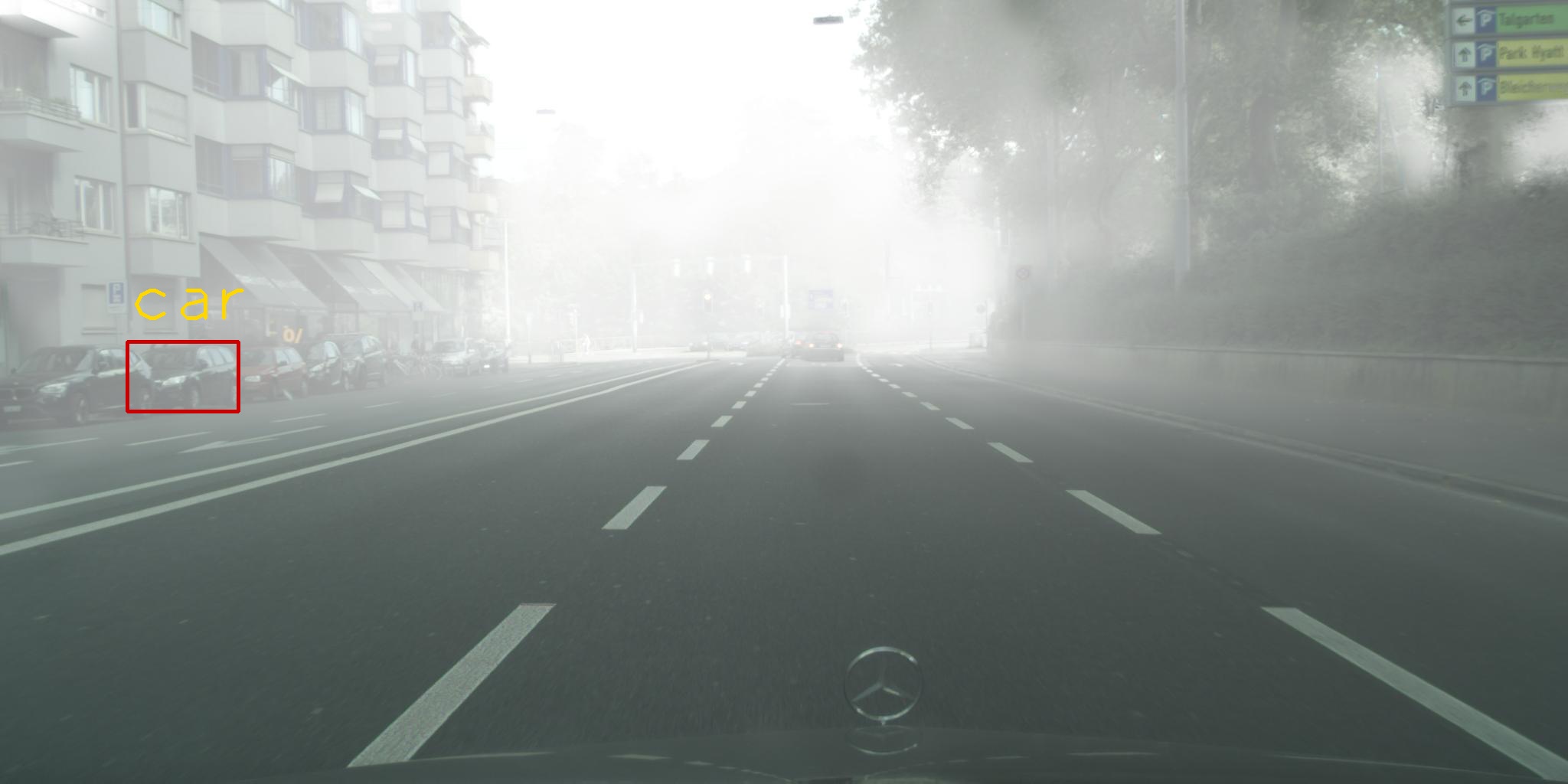}}
		\subfigure[FSANet]{\includegraphics[width=0.3\linewidth]{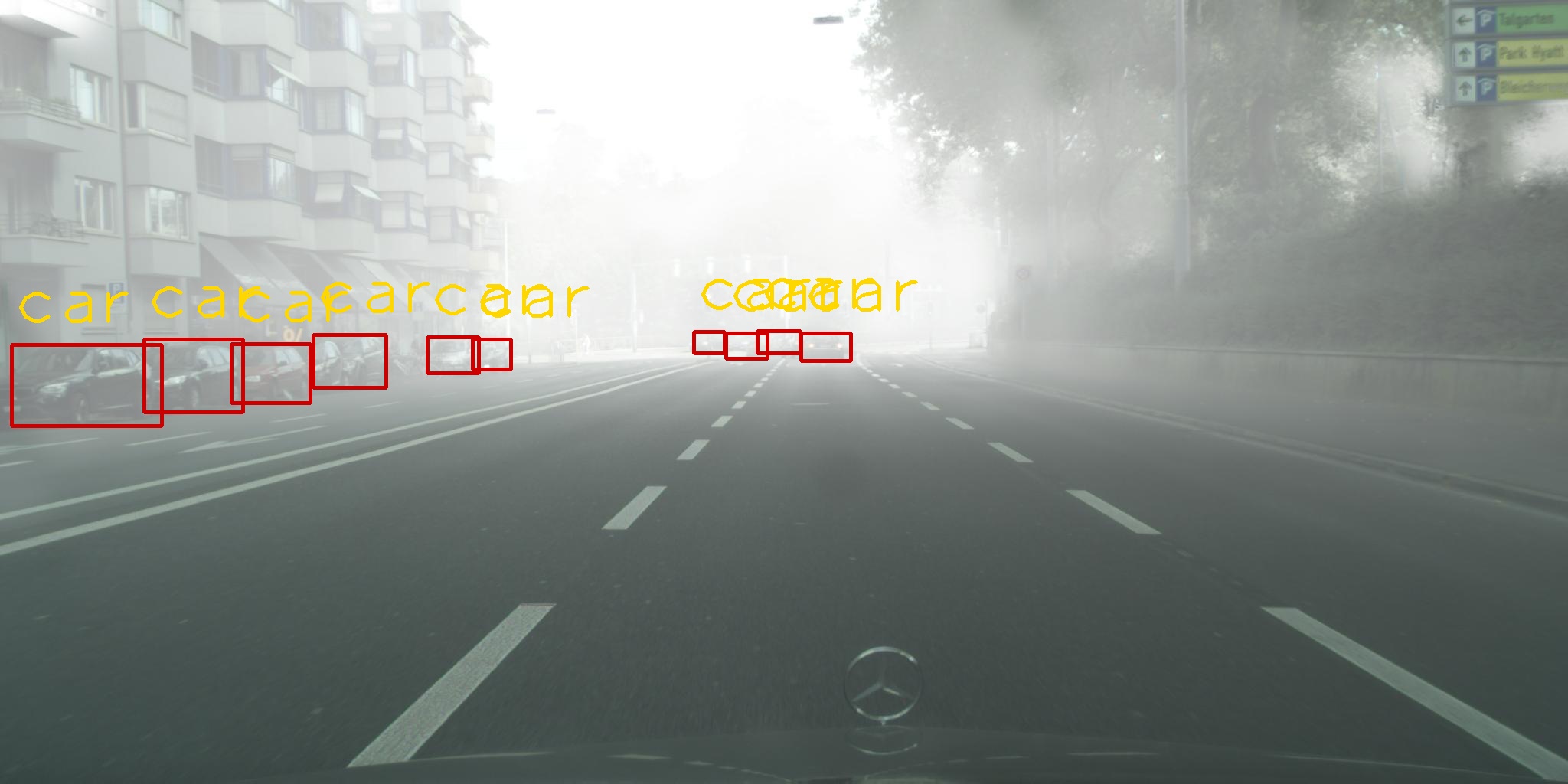}}\vspace{-2.5mm}\\
		\subfigure[Clipart1k]{\includegraphics[width=0.3\linewidth]{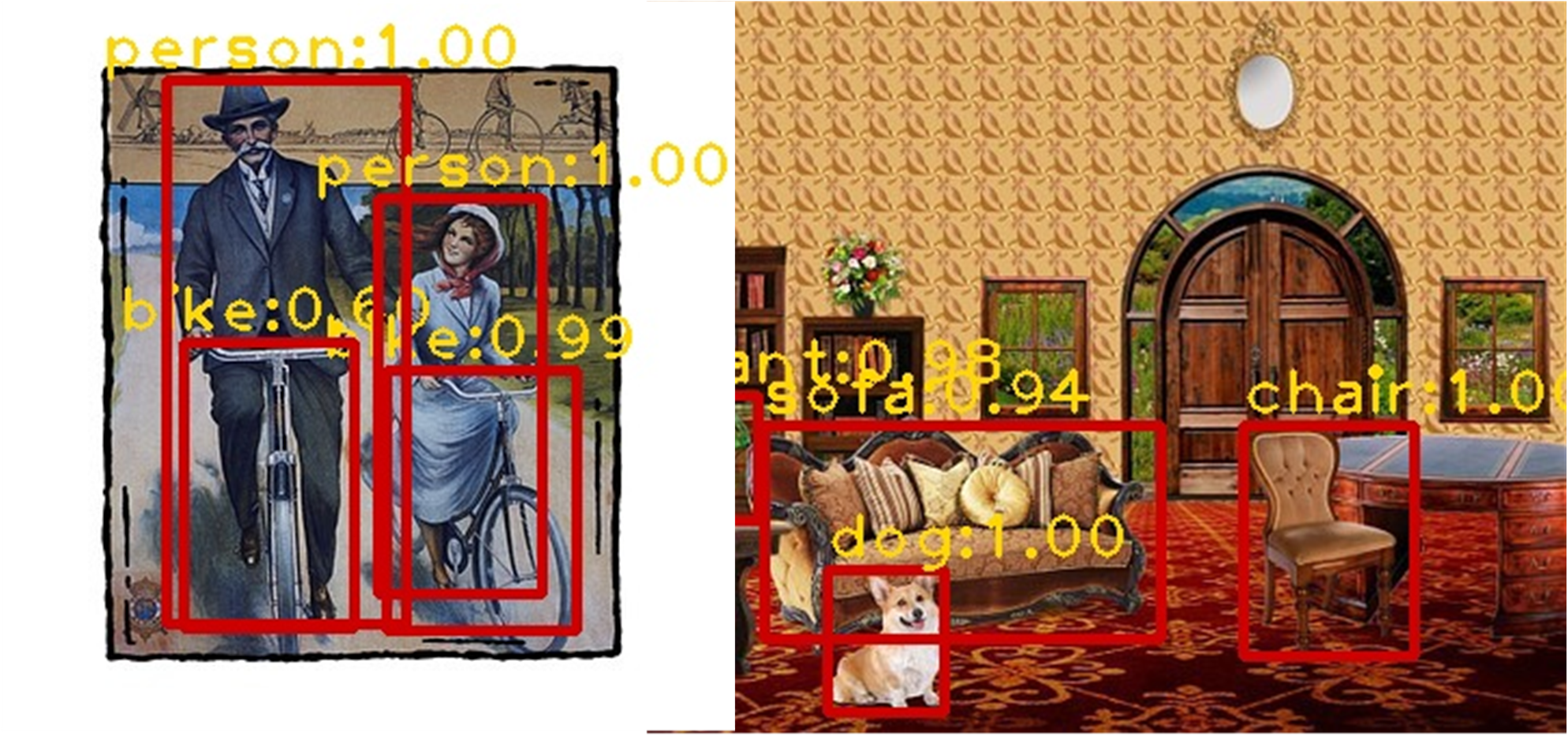}}
		\subfigure[Cityscapes]{\includegraphics[width=0.6\linewidth]{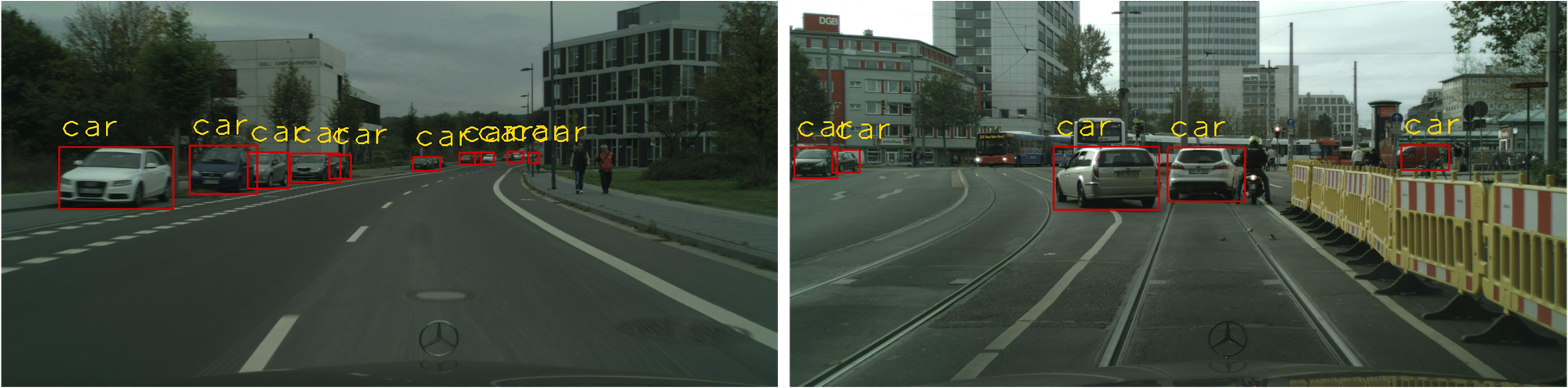}}
	\end{center}
	\caption{Examples of detection results on the target domain. From first to second rows (a): there is not foggy noise in the original images, (b-c): Cityscapes $\to$  Foggy-Cityscapes. (d): PASCAL $\to$ Clipart1k. (e): Sim10k $\to$ Cityscapes.}
	\label{fig:demo}
\end{figure*}
\begin{table*}[t]
	\setlength{\tabcolsep}{1pt}
	\begin{center}
		\resizebox{\textwidth}{!}{
			\begin{tabular}{lccccccccccccccccccccc}
				\toprule
				Method  & aero  & bike  & bird  & boat & bot & bus & car & cat & chair & cow & table & dog & horse & mbike & persn & plant & sheep & sofa & train & tv & mAP \\
				\midrule
				Source Only             & 35.6  & 52.5  & 24.3  & 23.0  & 20.0  & 43.9  & 32.8  & 10.7  & 30.6  & 11.7  & 13.8  & 6.0   & 36.8  & 45.9  & 48.7  & 41.9  & 16.5  & 7.3   & 22.9  & 32.0  & 27.8  \\
				Only LGFA                & 25.1  & 45.5  & 26.1  & 26.3  & 37.8  & 48.7  & 39.5  & 13.8  & 34.3  & 43.1  & \textbf{22.4}  & 11.2  & 36.5  & 62.4  & 57.1  & 50.4  & 16.0  & 26.4  & 48.9  & 42.3  & 35.7  \\
				FSANet w/o CSFS & \textbf{33.6} & 60.9  & 33.5  & \textbf{30.9}  & \textbf{43.7}  & 56.3  & 39.0  & \textbf{20.7}  & 35.5  & 58.2  & 13.7  & 22.4  & 35.6  & 81.2  & 60.8  & 48.0  & \textbf{28.2}  & 20.2  & 51.8  & 41.9  & 40.8  \\
				FSANet w/o RILA         & 32.5  & 62.2  & 32.4  & 30.5  & 42.3  & 53.6  & \textbf{42.8}  & 17.0  & 38.3  & \textbf{62.4}  & 20.2  & 19.9  & \textbf{36.9}  & 79.7  & 62.3  & 48.9  & 19.0  & 24.4  & 53.2  & 41.7  & 41.0  \\
				FSANet w/o LGFA          & 29.8  & 40.2  & 28.6  & 22.7  & 32.0  & 51.2  & 35.6  & 12.9  & 34.7  & 17.2  & 19.8  & 12.1  & 33.5  & 43.0  & 42.5  & 45.1  & 10.3  & \textbf{27.9}  & 42.9  & 40.9  & 31.1  \\
				FSANet w/o diff loss    & 32.8  & 60.4  & 32.1  & 30.5  & 38.7  & 70.6  & 40.0  & 19.9  & 37.2  & 56.4  & 22.1  & 22.0  & 34.9  & 71.9  & \textbf{63.9}  & 46.6  & 23.8  & 27.2  & 52.5  & 45.0  & 41.4  \\
				FSANet (RGB S)  & 26.8  & 53.4  & 33.0  & 30.6  & 37.1  & \textbf{72.3}  & 40.2  & 19.6  & \textbf{40.0}  & 59.3  & 21.6  & 13.9  & 32.8  & \textbf{85.7}  & 61.0  & \textbf{50.6}  & 19.7  & 26.1  & 53.3  & \textbf{45.6}  & 41.1  \\
				\midrule
				FSANet          & 31.0  &\textbf{63.7}  & \textbf{34.8}  & 29.4  & 43.0  & 70.7  & 40.8  & 18.7  & 39.6  & 57.4  & 22.2  & \textbf{27.0}  & 33.3  & 85.6  & 63.3  & 45.7  & 21.9  & 24.7  & \textbf{56.7}  & 44.5  & \textbf{42.7}  \\
				\bottomrule
			\end{tabular}
		}
	\end{center}
	\caption{Ablation experiments of FSANet on PASCAL $\to$ Clipart1k. More results on different datasets are shown in supplemental material.}
	\label{Ablation}
\end{table*}
\begin{table}[t]
	\begin{center}
		\begin{tabular}{lclc}
			\toprule
			Method          &      mAP      & Method          & mAP  \\ 
			\midrule
			Source Only     &     27.8      & K-Means ($K=2$) & 35.6 \\
			K-Means ($K=4$) &     41.6      & K-Means ($K=8$) & 39.8 \\
			\midrule
			SSF             & \textbf{42.7} &                 &      \\ 
			\bottomrule
		\end{tabular}
	\end{center}	
	\caption{Results of domain adaptation for object detection from PASCAL VOC to Clipart Dataset with K-Means or SSF.  The $K$ is the numbers of clustering of K-Means.}
	\label{kmeans}
\end{table}

\begin{figure}[!]
	\begin{center}
		\includegraphics[width=0.6\linewidth]{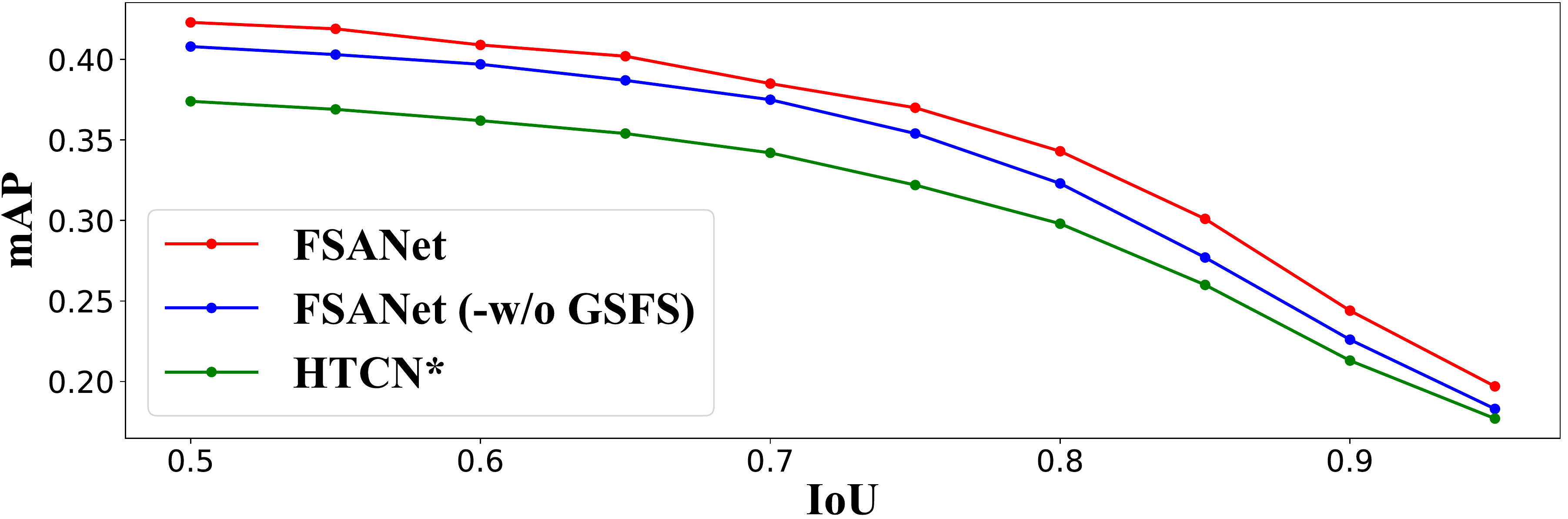}
	\end{center}
	\caption{The mAP with the variation of IoU thresholds on transfer task PASCAL$\to$Clipart1k. $\mathrm{HTCN}^*$ stands for HTCN trained without interpolation (w/o pixel-level alignment via CycleGAN~\cite{Chen2020Harmonizing}).}
	\label{fig:instance}
\end{figure}
\begin{figure*}[!]
	\begin{center}
		\includegraphics[width=0.8\linewidth]{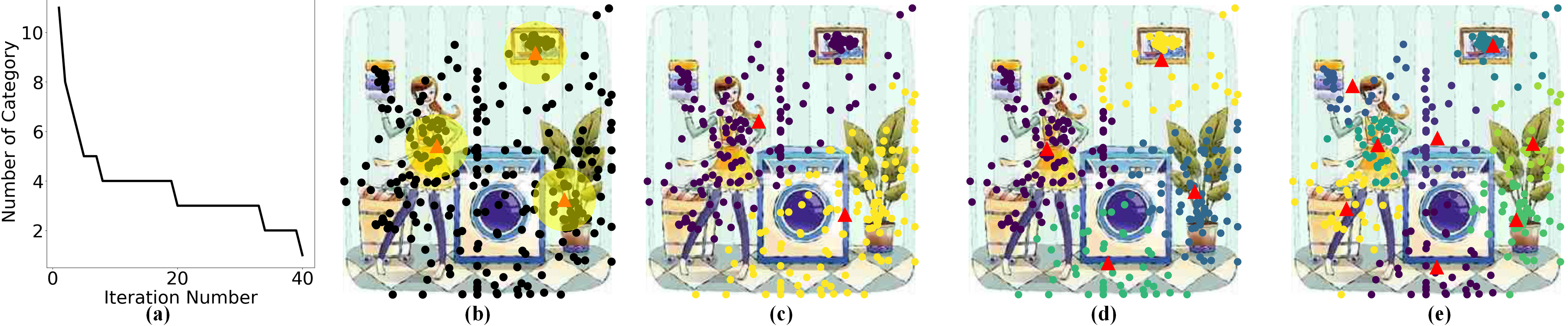}
	\end{center}
	\caption{Samples in PASCAL VOC $\to$ Clipart. 
		(a): The change curve of the ``Lifetime'' with the number of iterations. It can be seen that the ``Lifetime'' is the longest when $K=3$, so the number of adaptive categories is 3.
		(b): Predicting results of cluster centers and scales by SSF. The yellow circled area indicates the cluster center that is used for clustering, and the points outside the circle  will be deleted as outliers. Obviously, the category center basically coincides with the regions where the objects are located.
		(c-e): The clustering results of K-Means method in different $K$. From left to right: $K$=2, 4 and 8, respectively. Compared with the clustering result of SSF, the clustering centers of K-Means are away from the object regions.}
	\label{fig:kmeans}
\end{figure*}	
\begin{figure}[!]
	\begin{center}
		\includegraphics[width=0.65\linewidth]{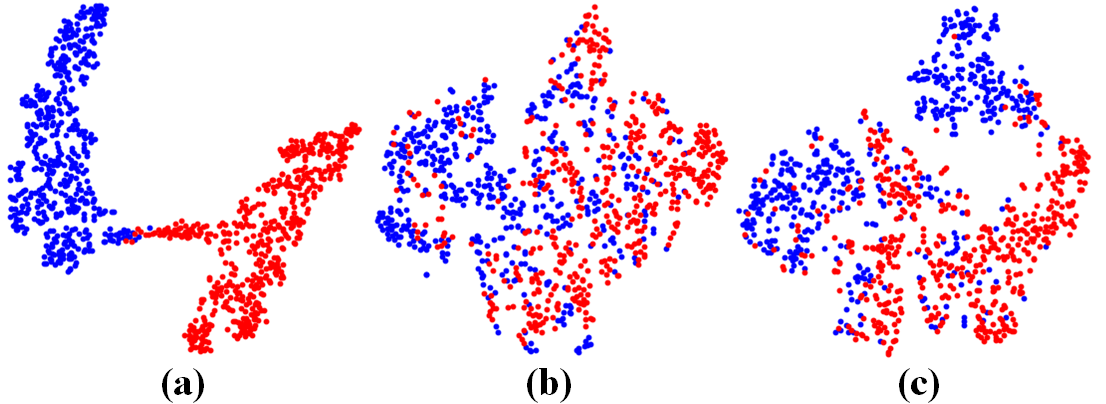}
	\end{center}
	\caption{Visualization of features obtained by transfer task Cityscapes $\to$ Foggy-Cityscapes. Blue/Red dots: source/target samples. (a-b): The distractive and shared features by FSANet, (c): The shared features by FSANet (w/o GSFS).}
	\label{fig:tsne}
\end{figure}

\subsection{Further Empirical Analysis}
In this section, some ablation experiments are established to show the rationality for different components in FSANet. Furthermore, we contrast the effectiveness between RILA and traditional instance-level alignment module and explain the advantage of SSF in RILA compared with other different clustering methods.

\noindent
{\bf Example of Detection Results. }
Examples of various adaptive object detection tasks on the target domain are exhibited in Figure~\ref{fig:demo}. 
We find that FSANet has good transferability and adaptivity on different datasets. 
For instance, from normal to foggy, FSANet can detect the indiscoverable objects in the foggy noise. 
As for the detection of artistic images, we can also get wonderful object and accurate bounding box predictions.

\noindent
{\bf Ablation Study.} We design various ablation experiments to prove the contributions of different modules in FSANet. They are shown as follows: {Only LGFA} denotes that we remove the RILA module and the Feature Separation module. {FSANet w/o GSFS} removes the GSFS module. {FSANet w/o RILA} eliminates the RILA module. {FSANet w/o LGFA} removes the LGFA module. {FSANet w/o diff loss} denotes that we do not use different loss in the GSFS module. {FSANet (RGB S)} denotes that we train the feature separation module by RGB images instead of gray-scale images.

\indent The results are reported in Table~\ref{Ablation} on PASCAL $\to$ Clipart1k. Compared with Only LGFA, adding RILA module or feature separation module can improve the performance of the model. 
FSANet w/o GSFS increases from $35.7\%$ to $40.8\%$ and the FSANet w/o RILA increases by $+5.3\%$. 
As shown in Figure~\ref{fig:tsne}, the GSFS module effectively promotes the features alignment. In addition, the separated distractive information is highly differentiated, demonstrating the rationality of feature separation.
While, mAP is only $31.1\%$ in the FSANet w/o LGFA, indicating that high-dimensional feature alignment is essential to avoid excessive separation of high-dimensional features.
Then, network performance will decline if different loss is not utilized, indicating that different loss is necessary to limit feature separation component.
Moreover, experiment FSANet (RGB S) shows that RGB images will reduce the performance of feature separation module.
In short, these experimental results fully demonstrate that the proposed module is more conducive to improving the transferability of the model.\\
\noindent
{\bf Region-Instance-level $\boldsymbol{v}$.$\boldsymbol{s}$. Instance-level. }
The mAP of different methods with the variation of IoU thresholds is displayed in Figure~\ref{fig:instance}.
Noticeably, the mAP reduces as the IoU threshold decreases and that of RILA is significantly improved compared with instance-level alignment on the IoU range of 0.50-0.95, which indicates that RILA can promote the adaptability and make bounding box regression more robustly.\\
{\bf Comparisons with K-Means.} We compare K-Means with our SSF algorithm to test whether the adaptively selecting region is efficient. In Table~\ref{kmeans}, it is obvious that mAP has huge fluctuations when $K$ is changed. In addition, in Figure~\ref{fig:kmeans}, the fixed category number reduces the flexibility for region selection, which shows the superiority for adaptive determination of region categories.

\section{Conclusion} 
In this paper, we propose a feature separation and alignment network (FSANet) for domain adaptive object detection. The FSANet can effectively decompose distractive information which is useless for detection by a feature separation module, and can restrain background noise and redundancy information by a region-instance-level alignment module, which can adaptively extract the regions to be aligned. Compared with existing methods, our novel FSANet can separate the distractive features and make our model focus on the features useful for detection. At the same time, region proposal redundancy and the background noise in feature alignment can be avoided.
Extensive experiments demonstrate that our method surpasses other existing models for adaptive object detection and the modules we proposed greatly improve the transferability and adaptability on several benchmark datasets.

	{\small
		\bibliographystyle{ieee_fullname}
		\bibliography{egbib}

\begin{thebibliography}{10}\itemsep=-1pt

\bibitem{Arruda2019Cross}
Vinicius~F Arruda, Thiago~M Paixão, Rodrigo~F Berriel, Alberto~F De~Souza,
  Claudine Badue, Nicu Sebe, and Thiago Oliveira-Santos.
\newblock Cross-domain car detection using unsupervised image-to-image
  translation: From day to night.
\newblock {\em International Joint Conference on Neural Networks}, pages 1--8,
  2019.

\bibitem{bousmalis2016domain}
Konstantinos Bousmalis, George Trigeorgis, Nathan Silberman, Dilip Krishnan,
  and Dumitru Erhan.
\newblock Domain separation networks.
\newblock In {\em Advances in neural information processing systems (NeurIPS)},
  pages 343--351, 2016.

\bibitem{cai2019exploring}
Qi Cai, Yingwei Pan, Chong-Wah Ngo, Xinmei Tian, Lingyu Duan, and Ting Yao.
\newblock Exploring object relation in mean teacher for cross-domain detection.
\newblock In {\em IEEE Conference on Computer Vision and Pattern Recognition
  (CVPR)}, pages 11457--11466, 2019.

\bibitem{Chen2020Harmonizing}
Chaoqi Chen, Zebiao Zheng, Xinghao Ding, Yue Huang, and Qi Dou.
\newblock Harmonizing transferability and discriminability for adapting object
  detectors.
\newblock In {\em IEEE Conference on Computer Vision and Pattern Recognition
  (CVPR)}, 2020.

\bibitem{DAfaster}
Yuhua Chen, Wen Li, Christos Sakaridis, Dengxin Dai, and Luc Van~Gool.
\newblock Domain adaptive faster r-cnn for object detection in the wild.
\newblock In {\em IEEE conference on computer vision and pattern recognition
  (CVPR)}, pages 3339--3348, 2018.

\bibitem{cordts2016cityscapes}
Marius Cordts, Mohamed Omran, Sebastian Ramos, Timo Rehfeld, Markus Enzweiler,
  Rodrigo Benenson, Uwe Franke, Stefan Roth, and Bernt Schiele.
\newblock The cityscapes dataset for semantic urban scene understanding.
\newblock In {\em IEEE conference on computer vision and pattern recognition
  (CVPR)}, pages 3213--3223, 2016.

\bibitem{deng2009imagenet}
Jia Deng, Wei Dong, Richard Socher, Li-Jia Li, Kai Li, and Li Fei-Fei.
\newblock Imagenet: A large-scale hierarchical image database.
\newblock In {\em IEEE conference on computer vision and pattern recognition
  (CVPR)}, pages 248--255. Ieee, 2009.

\bibitem{everingham2010pascal}
Mark Everingham, Luc Van~Gool, Christopher~KI Williams, John Winn, and Andrew
  Zisserman.
\newblock The pascal visual object classes (voc) challenge.
\newblock {\em International journal of computer vision}, 88(2):303--338, 2010.

\bibitem{ganin2015unsupervised}
Yaroslav Ganin and Victor Lempitsky.
\newblock Unsupervised domain adaptation by backpropagation.
\newblock In {\em International conference on machine learning}, pages
  1180--1189. PMLR, 2015.

\bibitem{girshick2015fast}
Ross Girshick.
\newblock Fast r-cnn.
\newblock In {\em IEEE international conference on computer vision (ICCV)},
  pages 1440--1448, 2015.

\bibitem{girshick2014rich}
Ross Girshick, Jeff Donahue, Trevor Darrell, and Jitendra Malik.
\newblock Rich feature hierarchies for accurate object detection and semantic
  segmentation.
\newblock In {\em IEEE conference on computer vision and pattern recognition
  (CVPR)}, pages 580--587, 2014.

\bibitem{goodfellow2014generative}
Ian Goodfellow, Jean Pouget-Abadie, Mehdi Mirza, Bing Xu, David Warde-Farley,
  Sherjil Ozair, Aaron Courville, and Yoshua Bengio.
\newblock Generative adversarial nets.
\newblock In {\em Advances in neural information processing systems (NeurIPS)},
  pages 2672--2680, 2014.

\bibitem{he2017mask}
Kaiming He, Georgia Gkioxari, Piotr Doll{\'a}r, and Ross Girshick.
\newblock Mask r-cnn.
\newblock In {\em IEEE international conference on computer vision (CVPR)},
  pages 2961--2969, 2017.

\bibitem{he2015spatial}
Kaiming He, Xiangyu Zhang, Shaoqing Ren, and Jian Sun.
\newblock Spatial pyramid pooling in deep convolutional networks for visual
  recognition.
\newblock {\em IEEE transactions on pattern analysis and machine intelligence},
  37(9):1904--1916, 2015.

\bibitem{he2016deep}
Kaiming He, Xiangyu Zhang, Shaoqing Ren, and Jian Sun.
\newblock Deep residual learning for image recognition.
\newblock In {\em IEEE conference on computer vision and pattern recognition
  (CVPR)}, pages 770--778, 2016.

\bibitem{he2019multi}
Zhenwei He and Lei Zhang.
\newblock Multi-adversarial faster-rcnn for unrestricted object detection.
\newblock In {\em IEEE International Conference on Computer Vision (CVPR)},
  pages 6668--6677, 2019.

\bibitem{he2020domain}
Zhenwei He and Lei Zhang.
\newblock Domain adaptive object detection via asymmetric tri-way faster-rcnn.
\newblock {\em arXiv preprint arXiv:2007.01571}, 2020.

\bibitem{Inoue2018Cross}
Naoto Inoue, Ryosuke Furuta, Toshihiko Yamasaki, and Kiyoharu Aizawa.
\newblock Cross-domain weakly-supervised object detection through progressive
  domain adaptation.
\newblock In {\em IEEE conference on computer vision and pattern recognition
  (CVPR)}, pages 5001--5009, 2018.

\bibitem{johnson2016driving}
Matthew Johnson-Roberson, Charles Barto, Rounak Mehta, Sharath~Nittur Sridhar,
  Karl Rosaen, and Ram Vasudevan.
\newblock Driving in the matrix: Can virtual worlds replace human-generated
  annotations for real world tasks?
\newblock {\em arXiv preprint arXiv:1610.01983}, 2016.

\bibitem{kim2019diversify}
Taekyung Kim, Minki Jeong, Seunghyeon Kim, Seokeon Choi, and Changick Kim.
\newblock Diversify and match: A domain adaptive representation learning
  paradigm for object detection.
\newblock In {\em IEEE Conference on Computer Vision and Pattern Recognition
  (CVPR)}, pages 12456--12465, 2019.

\bibitem{NIPS2012_4824}
Alex Krizhevsky, Ilya Sutskever, and Geoffrey~E Hinton.
\newblock Imagenet classification with deep convolutional neural networks.
\newblock In F. Pereira, C.~J.~C. Burges, L. Bottou, and K.~Q. Weinberger,
  editors, {\em Advances in Neural Information Processing Systems (NeurIPS)},
  pages 1097--1105. 2012.

\bibitem{li2020deep}
Wanyi Li, Fuyu Li, Yongkang Luo, and Peng Wang.
\newblock Deep domain adaptive object detection: a survey.
\newblock {\em arXiv preprint arXiv:2002.06797}, 2020.

\bibitem{lin2017focal}
Tsung-Yi Lin, Priya Goyal, Ross Girshick, Kaiming He, and Piotr Doll{\'a}r.
\newblock Focal loss for dense object detection.
\newblock In {\em IEEE international conference on computer vision (CVPR)},
  pages 2980--2988, 2017.

\bibitem{lin2014microsoft}
Tsung-Yi Lin, Michael Maire, Serge Belongie, James Hays, Pietro Perona, Deva
  Ramanan, Piotr Doll{\'a}r, and C~Lawrence Zitnick.
\newblock Microsoft coco: Common objects in context.
\newblock In {\em European conference on computer vision (ECCV)}, pages
  740--755. Springer, 2014.

\bibitem{liu2016ssd}
Wei Liu, Dragomir Anguelov, Dumitru Erhan, Christian Szegedy, Scott Reed,
  Cheng-Yang Fu, and Alexander~C Berg.
\newblock Ssd: Single shot multibox detector.
\newblock In {\em European conference on computer vision (ECCV)}, pages 21--37.
  Springer, 2016.

\bibitem{Long2017Fully}
Long, Jonathan, Shelhamer, Evan, Darrell, and Trevor.
\newblock Fully convolutional networks for semantic segmentation.
\newblock {\em IEEE transactions on pattern analysis and machine intelligence},
  39(4):640--651, 2017.

\bibitem{paszke2017automatic}
Adam Paszke, Sam Gross, Soumith Chintala, Gregory Chanan, Edward Yang, Zachary
  DeVito, Zeming Lin, Alban Desmaison, Luca Antiga, and Adam Lerer.
\newblock Automatic differentiation in pytorch.
\newblock 2017.

\bibitem{Redmon_2016_CVPR}
Joseph Redmon, Santosh Divvala, Ross Girshick, and Ali Farhadi.
\newblock You only look once: Unified, real-time object detection.
\newblock In {\em IEEE Conference on Computer Vision and Pattern Recognition
  (CVPR)}, pages 779--788, June 2016.

\bibitem{Redmon_2017_CVPR}
Joseph Redmon and Ali Farhadi.
\newblock Yolo9000: Better, faster, stronger.
\newblock In {\em IEEE Conference on Computer Vision and Pattern Recognition
  (CVPR)}, pages 6517--6525, July 2017.

\bibitem{redmon2018yolov3}
Joseph Redmon and Ali Farhadi.
\newblock Yolov3: An incremental improvement.
\newblock {\em arXiv preprint arXiv:1804.02767}, 2018.

\bibitem{ren2015faster}
Shaoqing Ren, Kaiming He, Ross Girshick, and Jian Sun.
\newblock Faster r-cnn: Towards real-time object detection with region proposal
  networks.
\newblock In {\em Advances in neural information processing systems (NeurIPS)},
  pages 91--99, 2015.

\bibitem{saito2019strong}
Kuniaki Saito, Yoshitaka Ushiku, Tatsuya Harada, and Kate Saenko.
\newblock Strong-weak distribution alignment for adaptive object detection.
\newblock In {\em IEEE Conference on Computer Vision and Pattern Recognition
  (CVPR)}, pages 6956--6965, 2019.

\bibitem{sakaridis2018semantic}
Christos Sakaridis, Dengxin Dai, and Luc Van~Gool.
\newblock Semantic foggy scene understanding with synthetic data.
\newblock {\em International Journal of Computer Vision}, 126(9):973--992,
  2018.

\bibitem{simonyan2014very}
Karen Simonyan and Andrew Zisserman.
\newblock Very deep convolutional networks for large-scale image recognition.
\newblock {\em arXiv preprint arXiv:1409.1556}, 2014.

\bibitem{sindagi2019prior}
Vishwanath~A. Sindagi, Poojan Oza, Rajeev Yasarla, and Vishal~M. Patel.
\newblock Prior-based domain adaptive object detection for hazy and rainy
  conditions.
\newblock In {\em European conference on computer vision (ECCV)}, volume 12359
  of {\em Lecture Notes in Computer Science}, pages 763--780. Springer, 2020.

\bibitem{uijlings2013selective}
Jasper~RR Uijlings, Koen~EA Van De~Sande, Theo Gevers, and Arnold~WM Smeulders.
\newblock Selective search for object recognition.
\newblock {\em International journal of computer vision}, 104(2):154--171,
  2013.

\bibitem{895974}
{Yee Leung}, {Jiang-She Zhang}, and {Zong-Ben Xu}.
\newblock Clustering by scale-space filtering.
\newblock {\em IEEE Transactions on Pattern Analysis and Machine Intelligence},
  22(12):1396--1410, 2000.

\bibitem{zhaoijcai2020}
Zixiang Zhao, Shuang Xu, Chunxia Zhang, Junmin Liu, Jiangshe Zhang, and Pengfei
  Li.
\newblock {DIDFuse}: Deep image decomposition for infrared and visible image
  fusion.
\newblock In {\em International Joint Conference on Artificial Intelligence
  (IJCAI)}, pages 970--976, 2020.

\bibitem{Zhu2017Unpaired}
Jun{-}Yan Zhu, Taesung Park, Phillip Isola, and Alexei~A. Efros.
\newblock Unpaired image-to-image translation using cycle-consistent
  adversarial networks.
\newblock In {\em {IEEE} International Conference on Computer Vision (ICCV)},
  pages 2242--2251, 2017.

\bibitem{Xinge2019Adapting}
Xinge Zhu, Jiangmiao Pang, Ceyuan Yang, Jianping Shi, and Dahua Lin.
\newblock Adapting object detectors via selective cross-domain alignment.
\newblock In {\em IEEE Conference on Computer Vision and Pattern Recognition
  (CVPR)}, pages 687--696, 2019.

\end{thebibliography}
	}
	
\end{document}